\documentclass[letterpaper]{article} % DO NOT CHANGE THIS
\usepackage[]{aaai23}  % DO NOT CHANGE THIS
\usepackage{times}  % DO NOT CHANGE THIS
\usepackage{helvet}  % DO NOT CHANGE THIS
\usepackage{courier}  % DO NOT CHANGE THIS
\usepackage[hyphens]{url}  % DO NOT CHANGE THIS
\usepackage{graphicx} % DO NOT CHANGE THIS
\usepackage{natbib}  % DO NOT CHANGE THIS AND DO NOT ADD ANY OPTIONS TO IT
\usepackage{caption} % DO NOT CHANGE THIS AND DO NOT ADD ANY OPTIONS TO IT
\usepackage{algorithm}
\usepackage{algorithmic}

\newcommand{\comment}[1]{}

\usepackage{csquotes}
\MakeOuterQuote{"}

\usepackage{newfloat}
\usepackage{listings}
\DeclareCaptionStyle{ruled}{labelfont=normalfont,labelsep=colon,strut=off} % DO NOT CHANGE THIS
\floatstyle{ruled}
\newfloat{listing}{tb}{lst}{}
\floatname{listing}{Listing}
\title{Multimodal Propaganda Processing}
\author{
    Vincent Ng and
    Shengjie Li
}
\affiliations{
    Human Language Technology Research Institute\\
    University of Texas at Dallas\\
    Richardson, TX 75083-0688\\
    \{vince,sxl180006\}@hlt.utdallas.edu
}

\usepackage{bibentry}
\begin{document}

% \includepdf[pages=-]{abstract2}
% \includepdf[pages=-]{bio6}

\maketitle

\begin{abstract}
Propaganda campaigns have long been used to influence public opinion via disseminating biased and/or misleading information.
%achieve personal agendas such as influencing public opinions. 
Despite the increasing prevalence of propaganda content on the Internet, few attempts have been made by AI researchers to analyze such content. We introduce the task of multimodal propaganda processing, where the goal is to automatically analyze propaganda content.
%by extracting relevant information from the different modalities present in the content, identifying the persuasion devices used, and eventually generating the hidden message being conveyed. 
We believe that this task presents a long-term challenge to AI researchers and that successful processing of propaganda could bring machine understanding one important step closer to human understanding. We discuss the technical challenges associated with this task and outline the 
%initial 
steps that need to be taken to address it. 
%Given that NLP and CV researchers have focused largely on analyzing information that is explicitly expressed in text and images, we believe that the need to uncover hidden agendas by combining information extracted from multiple input modalities as well as external background information makes this task an ideal long-term challenge for AI researchers.

\end{abstract}

\begin{figure*}[t]
    \centering
    \includegraphics[width=.73\textwidth]{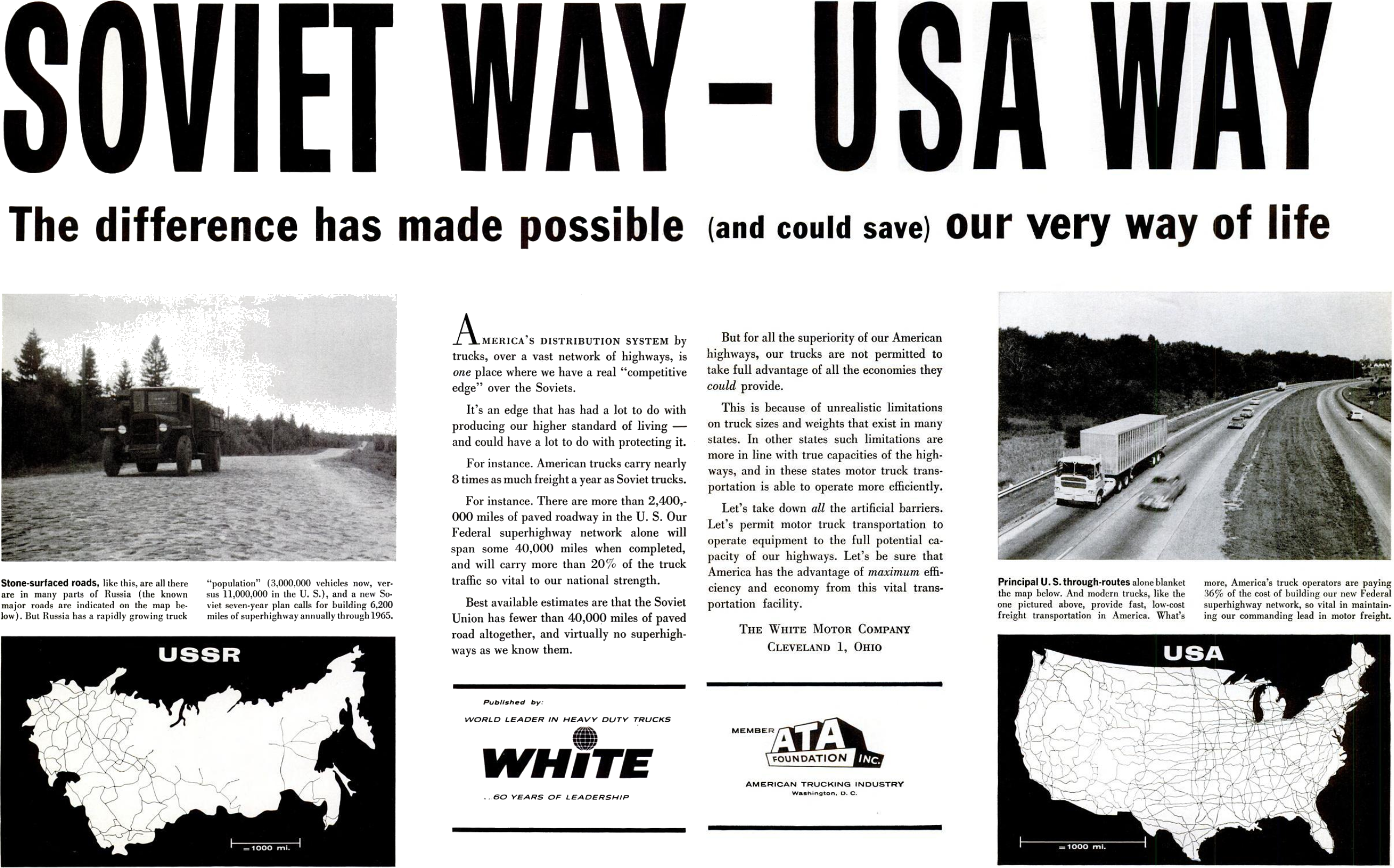}
    \caption{An advertisement published by the White Motor Company in 1965.}
    \label{fig:ussr-us}
    \vspace{-2mm}
\end{figure*}

\section{Introduction}

%In the past two decades, 
Since the beginning of this century,
significant progress has been made in the area of Sentiment Analysis and Opinion Mining on  processing opinionated documents. Recent years have seen a surge of interest in processing a particular type of opinionated documents: persuasive documents. Work in this area is typically done under the umbrella of Argument Mining, in which the core task is to uncover the argumentative structure of a persuasive document. Specifically, the goal is to (1) identify the main claim, the claims, and the premises (i.e., supporting evidences) expressed in the given document, and (2) determine the relationships among them (e.g., identify which premises support which claim).
%which claims support the major claim).

Work on argument mining has so far focused on processing legal text \cite{legal, legal09, walker-etal-2018-evidence}, persuasive student essays \cite{persing-ng-2016-end,stab-gurevych-2017-parsing}, and Oxford-style debates \cite{orbach-etal-2020-echo, Slonim2021}.
Although persuasive in nature, propagandistic articles (i.e., articles that aim to influence public opinion via disseminating biased and/or misleading information)  have received relatively little attention in Natural Language Processing (NLP). This is somewhat surprising %considering 
given the growing prevalence of {\em computational propaganda}, an "emergent form of political manipulation that occurs over the Internet" \cite{book:computational18}.
%nowadays. 
From a research perspective, automatic processing of propaganda %propagandistic contents 
%are interesting in that the automatic processing of such contents 
presents a number of challenges to AI researchers:

%\vspace{-2mm}
\paragraph{Multimodality.} One characteristic that distinguishes propaganda content from other persuasive texts is that the former is often multimodal, comprising both text and images. As the saying goes, a picture is worth a thousand words. In multimodal propaganda, it is often the images that are most eye-catching and which create the biggest psychological impact on the reader. Although the text usually plays a supporting role, there are many cases where the image(s) could not be understood properly without the supporting text. How to combine the information derived from the two modalities to properly understand propaganda is an open question.

\vspace{-2mm}
\paragraph{Deep understanding of text and images.} Propaganda processing takes argument mining to the next level of complexity. As noted above, argument mining involves (1) extracting the claims and premises from the associated text and (2) identifying the relationships (e.g., support, attack) among them.
%claims and the premises. 
For the kind of texts that NLP researchers have focused on so far (e.g., legal text, %argumentative student essays), %and 
Oxford-style debates), %all 
the claims and premises are typically clearly stated. In contrast, the main claims and possibly some other supporting claims in propagandistic articles are often intentionally omitted, so we are faced with the additional challenge of recovering these hidden messages. Moreover, while the arguments in legal texts, essays, and debates can largely be interpreted literally, we often have to read between the lines when interpreting the text and images in propaganda content. For example, when given a picture of Russian soldiers killing Ukrainian civilians, current Computer Vision (CV) technologies would be able to produce a caption about this killing event, but if this picture appears in propagandistic articles, we probably need to infer the motive behind this picture (e.g., gaining the world's sympathy and support for Ukraine), which is currently beyond the reach of today's technology.
%something that current technologies are not yet capable of doing. 

\vspace{-2mm}
\paragraph{The need for background knowledge.} Historical or cultural background knowledge may be needed to properly process propaganda content. For instance, given a propagandistic article with a picture showing a Palestinian social unrest event in the West Bank, the author may want to instill fear among the Israelis. However, without the %historical 
knowledge of the long-standing conflict between the Palestinians and the Israelis, one may not be able to %properly 
understand the author's intent.

\vspace{-2mm}
\paragraph{Persuasion by deception.} As noted above, argument mining researchers have focused on processing legal text, essays, and debates, where virtually all claims are established using persuasion strategies like logos (i.e., through logical reasoning), pathos (i.e., through an emotional appeal), and ethos (i.e., through the speaker's authority or status). In contrast, the persuasion strategies used in propaganda are %a lot 
more sophisticated, often involving logical fallacies and framing.

Automatic processing of propaganda content could have important societal ramifications. In many cases, people are not aware that they are being brainwashed by propaganda campaigns, and this could lead to life-threatening consequences. One of the most compelling examples would be the ISIS propaganda and recruitment in 2014 \cite{farwell2014media}, in which ISIS successfully recruited many people from all over the world, particularly those from the European Union, to go to Syria to serve as soldiers and sex slaves.
Within the U.S.\, propaganda is typically manifested in the form of political manipulation campaigns with the goal of swaying public opinion. In fact, political manipulation campaigns have doubled since 2017 \cite{bradshaw2018challenging}, and increased efforts of disinformation should be expected as the U.S.\ midterm elections draw near \cite{newKnowledge}.

In this paper, we present the novel task of multimodal propaganda processing, where the goal is to analyze propaganda content by extracting relevant information from the different modalities present in the content, identifying the persuasion devices and tactics that are used in different portions of the content, and eventually generating the message(s) being conveyed. We believe that time is ripe for AI researchers to work on this task. From a societal perspective, given the increasing influence that propaganda content has on our daily lives, it is more important than ever for us to be able to understand propagandistic articles. From an AI perspective, deep learning technologies have enabled revolutionary advances in machine understanding. % in the last decade, 
It is time to examine how robust these technologies are when applied to a task as challenging as multimodal propaganda processing. 
%Since this task requires a deep understanding of text and images, the successful development of a multimodal propaganda processing system would imply that machine understanding is one step closer to human understanding.

\section{Examples}

In this section, we explain why multimodal propaganda processing is interesting and challenging via two examples.
% . We discuss two examples, where the first one conveys propaganda messaging mainly through text and the second one conveys propaganda messaging mainly through images. 
%Both of them are challenging but in different aspects.

\label{sec:en-guardia}

% \begin{figure*}[t]
%     \centering
%     \includegraphics[width=.7\textwidth]{tree.pdf}
%     \caption{Some pages of the \textit{En Guardia} magazine published by US propagandists to seek support from the Latin Americas.}
%     \label{fig:en-guardia}
% \end{figure*}

\subsection{Example 1}
\label{sec:ussr-us}

Figure \ref{fig:ussr-us} presents an advertisement published by the White Motor Company in 1965 aiming to establish the superiority of the American truck transportation road networks to their Soviet counterparts. The advertisement shows on the left side
%superiority of their trucks. 
%The advertisement %was originally on a magazine and it 
%used the spread of the magazine in which it appeared fairly well. 
%The left side of the advertisement shows how people in the USSR transported merchandise and other things, where 
the stone-surfaced roads being used for transportation in the USSR and the paucity of paved roads in the country, and it shows on the right side the
modern highways along with a map of the USA that is full of road networks.
%On the right side of the advertisement, modern highways in the US are shown along with a map of the US that is full of road networks. 
The text states that (1) the distribution system by trucks %in the US 
was only one of the examples where the USA was superior to the USSR,
%places that the Americans were over the Soviets, 
and (2) the restrictions imposed by many states in the USA concerning truck sizes prevented motor truck transportation from operating to its full potential.

\begin{figure}[t]
    \centering
    \includegraphics[width=.475\textwidth]{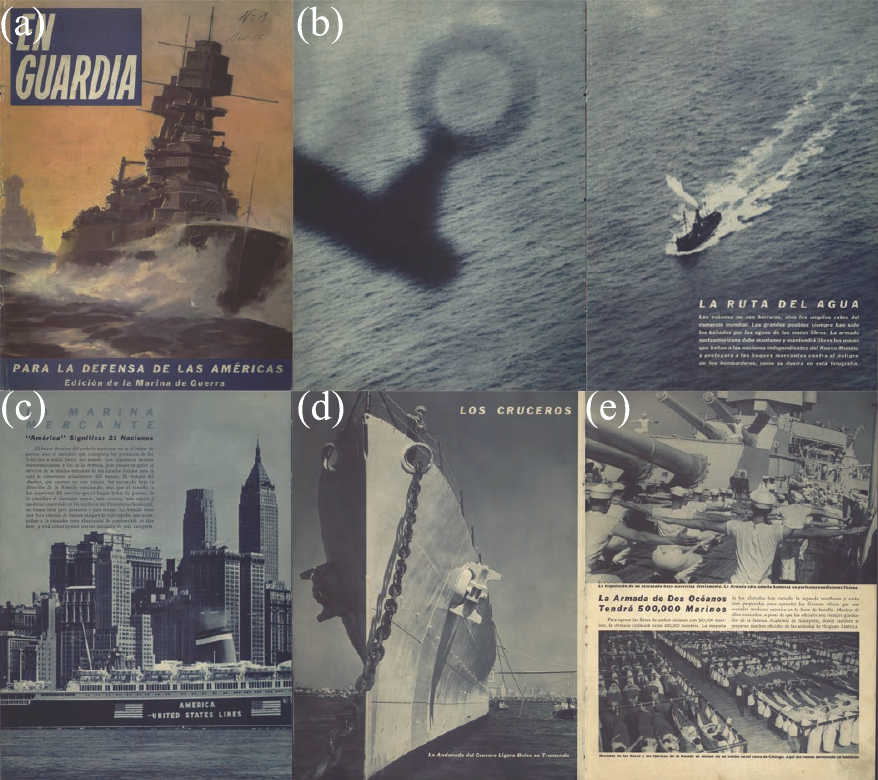}
    \caption{Some pages of the \textit{En Guardia} magazine.
    %published by the U.S.\ propagandists to seek support from the Latin Americas. 
    Figure~2(b) covers two pages.}
    \label{fig:en-guardia}
    \vspace{-2mm}
\end{figure}

\vspace{-2mm}
\paragraph{Human perception.} Human readers need to possess certain knowledge in order to process the propaganda content in this %piece of 
advertisement. First, they have to have some geology knowledge to discover that the two maps are at different scales, which make the USA look larger than the USSR in terms of land mass. Second, they need to be aware of the oversimplifying language "SOVIET WAY - USA WAY", which implies that the Soviets adopted the American way of transportation. 
No evidence was provided to substantiate this claim, however.
%while their main way of transportation could have been railway transportation. 
Third, they need to pay attention to the deceptive language in the text. While the advertisement contains a road map of the USSR and an estimated length of paved road in the country, the sources were never given and hence the information could be far from accurate. 

%Specifically, the following 
Several propaganda devices and tactics are involved in this 
%piece of 
advertisement
%\footnote{
(see Section~3 for the list of devices and tactics). First, 
%the readers will be first attracted by the title and the subtitle, as they have a large font size and take up a huge space of the entire page. 
the large font size associated with the title and the subtitle as well as the oversimplifying language in them signal the use of the \textit{Binary Reduction} %propaganda 
tactic, which employs the false-dilemma logical fallacy. Second, the use of language to depict that the Americans are superior in many ways (e.g., "Our higher standard of living", "for all the superiority of our American highways", etc.) signals the \textit{All-encompassing} %propaganda 
tactic, which is a sort of rhetoric that often appears as window dressing for a larger point. Third, the sentence "America has the advantage of maximum efficiency and economy" employs the \textit{Cultural Signaling} %propaganda 
tactic, calling on America's values of efficiency and success. Finally, the text gives "best estimates" of the USSR's road network, which is a case of the \textit{Card Stacking} %propaganda 
device, where only partial facts are used to defend a statement.

\vspace{-2mm}
\paragraph{Challenges.} 
It is non-trivial %for a computer system 
to automatically process the propaganda content in this advertisement.
The key challenges stem from the need to (1)
%involves a number of challenges. First, 
%it is %naturally 
process multimodal information extracted from the images and the text;
(2) %Second, 
%the computer system needs to have enough 
exploit background knowledge to unveil hidden information;
%which is not easy even for human readers outside of the targeted audience. 
%Third, 
(3) interpret the use of different font styles and text sizes to highlight specific pieces of information; and (4) understand the hidden information conveyed in the images (e.g., the difference in scaling between the two maps).
%the various font styles and sizes of text are 
%may be obvious for human readers but not so for computer systems. Fourth, hidden information in the images also poses a challenge to current computer vision systems. In this case, it might be easy for a human reader to notice the difference in scaling between the two maps, but it is not easy for a computer system to learn and encode such information.

\subsection{Example 2}

During World War II, the U.S.\ propagandists %in the US 
sought %to win 
support from the Latin Americas by publishing a high-quality Spanish periodical \textit{En Guardia}. Figure \ref{fig:en-guardia} shows six pages %taken 
from the first issue of %the 
\textit{En Guardia}.
%magazine.
%While 
Each %of the six pages 
page has its own propaganda messaging.
%conveyed by the author. 
Figure \ref{fig:en-guardia}(a) is the cover of the magazine, which shows two naval ships moving fast in the ocean. The title "En Guardia" and the subtitle "Para la defensa de las Am\'{e}ricas" translate into "On Guard" and "For the defense of the Americas" respectively. Figure \ref{fig:en-guardia}(b) %and \ref{fig:en-guardia}(c) 
shows a naval ship in the ocean with a scope pointing at the ship. The main points in the caption of \ref{fig:en-guardia}(b) translate into "The American navy must and will keep the seas 
% that bathe the independent nations of the New World 
free, and will protect merchant ships against the danger of bombardment". Figure \ref{fig:en-guardia}(c) shows a merchant ship and discusses the importance of merchant ships in delivering goods and troops to all parts of the Americas. The boldfaced sub-caption translates into "America means 21 nations". Figure \ref{fig:en-guardia}(d) shows a ship while Figure \ref{fig:en-guardia}(e) focuses on the training and the sheer size of the U.S.\ Navy. 
% Figure \ref{fig:en-guardia}(g) contains an image and an article describing the interactions of US naval personnel with a popular celebrity. Figure \ref{fig:en-guardia}(h) is an image of a midshipman with a date pictured in front of a cannon. 
Most importantly, these pages need to be considered as a {\em sequence} in order to obtain the full %propaganda 
messaging, which is that "Maritime commerce in the Americas is under imminent threat, and protecting the oceans from the enemy is vital to western hemisphere interests. However, the U.S. navy has the best equipment and personnel to deal with such a threat."

\vspace{-2mm}
\paragraph{Human perception.} A human reader with the appropriate background would interpret these pages as follows. First, the cover points out the main theme of the magazine by using the eye-catching subtitle "For the defense of the Americas", which proposes a sense of shared identity and immediate danger. The warship depicted on the cover is moving fast in the ocean, as the water below it is splashing high, suggesting something is happening immediately. Second, Figure \ref{fig:en-guardia}(b) shows an exaggerated scaling of a gun scope on the left and %while the right half of Figure \ref{fig:en-guardia}(b) shows 
a ship that is being pointed to by the scope on the right. This would naturally take all the attention of the reader. Third, Figure \ref{fig:en-guardia}(c) repeats the sense of a shared identity by saying "America means 21 nations". Finally, Figures \ref{fig:en-guardia}(d) and \ref{fig:en-guardia}(e) show that the U.S.\ has the equipment and the personnel to deal with the danger threatening free commerce. 
% Figures \ref{fig:en-guardia}(g) and \ref{fig:en-guardia}(h) show not only do the US marines have the capacity to deal with incoming threats, they are also treated well and can enjoy some leisure time. When these pages are considered in a sequence, the message the sequence is trying to convey pops out: the Americas as a whole are faced with immediate danger, and the US military will protect every nation with its military strength.

Next, we analyze the propaganda devices and tactics used in these pages. Figures \ref{fig:en-guardia}(a), \ref{fig:en-guardia}(b), and \ref{fig:en-guardia}(d) all use the \textit{Visual Scaling} %propaganda 
tactic, which is concerned with evoking emotional understanding (e.g., fear, power, etc.) by using images. These images also use the \textit{Card Stacking} %propaganda 
device, as they do not explicitly point out who is threatening the Americas. Figure \ref{fig:en-guardia}(c) uses two %propaganda 
devices: (1) \textit{Band Wagon}, which implies that all of the countries in the Western Hemisphere are a collective and should work together; and (2) \textit{Glittering Generalities}, where a "virtue word" (in this case, the impressiveness of U.S.\ merchant ships) is being used to create positive emotion and acceptance
(of the U.S.\ military involvement) 
without examination of evidence. 
% Figures \ref{fig:en-guardia}(g) and \ref{fig:en-guardia}(h) use the \textit{Plain Folks} propaganda device which attempts to convince the audience that the propaganda ideas are good because the audience and the creator are alike. 

\vspace{-2mm}
\paragraph{Challenges.} Automatically processing propaganda in this example is even harder than that in the first example 
%in Section \ref{sec:ussr-us} for a computer system 
since proper understanding depends heavily on visual clues rather than textual information. For instance, a machine needs to understand that (1) in Figure \ref{fig:en-guardia}(b), the scope was enlarged to an exaggerated size and was pointing at the ship; (2) in Figure \ref{fig:en-guardia}(d), the picture was taken at an angle that makes the ship look substantially larger than 
%huge compared to 
other objects in the background, with the intent of showing off %Besides the size itself, 
%In addition to size, the computer system needs to %understand that human readers will relate the massive-looking ship to 
the military might of the U.S.; and (3) %Furthermore, 
these images need to be considered as a {\em sequence} in order to get the full messaging.
%which adds another dimension of difficulty.
%This adds another dimension of difficulty as the sequence itself is hard to capture by a computer system.

\section{Corpus Creation and Annotation}
\label{sec:3}

In this section, we outline the initial steps 
%that we believe need to be taken in order to 
needed to address the task of multimodal propaganda processing.

\subsection{Corpus Creation}

Given the recent advances in CV and NLP, we propose to approach this problem by building a machine learning (ML) system. Appropriately annotated corpora are critical to the successful application of any ML systems.
Since the goal of the multimodal propaganda processing task is to analyze propaganda content, we need data instances that correspond to examples of propaganda.
Since we do not have a system for automatically identifying propaganda content, it would be best for us to begin data collection by looking for websites or publications that are known to publish propagandistic materials. A possible source of historical propaganda would be the \textit{En Guardia} magazine described in the previous section. So far, we have applied OCR to every page of every issue of this magazine and have used these articles to assemble the first version of our corpus.

In order to assemble a corpus that contains contemporary propagandistic articles, 
% we propose to perform a keyword-based search to automatically identify candidates of propaganda content, and 
we propose to exploit the content published on some fact-checking websites. For example, Politifact\footnote{https://www.politifact.com/} verifies the accuracy of claims made by elected officials. Those claims that were marked as inaccurate would constitute a good set of candidates of propagandistic articles. 
We can then manually go through these candidates to identify propaganda content. Similar websites include Full Fact\footnote{https://fullfact.org/}, FactCheck\footnote{https://www.factcheck.org/}, and Media Bias/Fact Check\footnote{https://mediabiasfactcheck.com/}.
% Our initial investigation reveals that useful keywords include xxx.

%and will use it as our first corpus for investigating this task.

%Corpora of different domains and different languages can be collected based on individual needs. For example, if interested in wartime propaganda in Spanish, the \textit{En Guardia} magazine that US propagandists published during World War II could be a great source. If interested in more recent political events, articles from general news agencies are a great source.

\begin{figure*}[t]
    \centering
    \includegraphics[width=.9\textwidth]{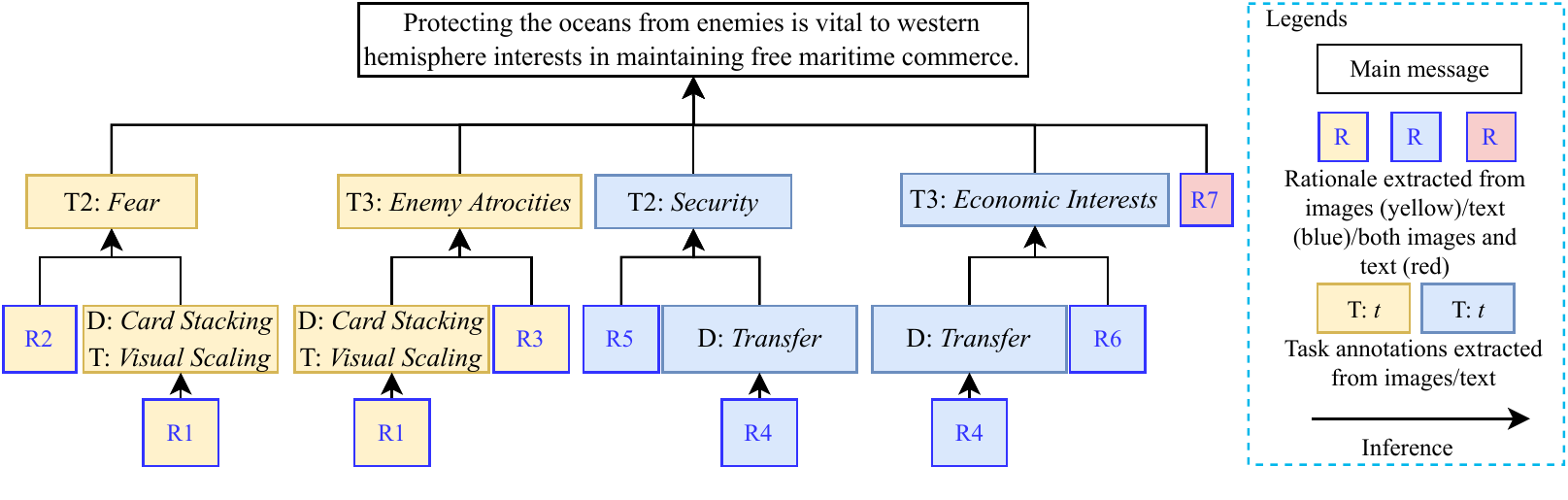}
    \caption{Sample argument tree for the input taken from Figure~2(b). Given this figure, a possible caption generated for Task 6 is "A ship is sailing on the sea, while a large gun scope is pointing at it". R1: A gun scope is pointing at a ship, which creates a sense of danger, hence \textit{Visual Scaling}. However, who is holding the gun is unclear (partial information), hence \textit{Card Stacking}. R2: A sense of \textit{Fear} created by the gun scope. R3: the image depicts a (potential) crime of bombardment, hence \textit{Enemy Atrocities}. R4: Use the idea of the U.S.\ protecting maritime commerce to justify other U.S.\ military involvements, hence \textit{Transfer}. R5: "La armada norteamericana ... proteger\'{a} a los buques mercantes" means "the American army will protect merchant vessels", which suggests a sense of  \textit{Security}. R6: "Los océanos no son barreras, sino las amplias rutas del comercio mundial" means "the oceans are the routes of world trades", which focuses on \textit{Economic Interests}. R7: Since the image suggests threats to maritime trades, while the text suggests the U.S.\ will protect maritime trades, we can get the overall message by combining them.}
    %Underlined items are the hidden information that needs to be extracted.}
    \label{fig:exp}
    \vspace{-2mm}
\end{figure*}

\subsection{Annotation Tasks}

Next, we define a set of annotation tasks that we believe would be helpful for analyzing propaganda content. The annotations we obtain via these tasks will provide the data needed to train models for processing propaganda content. 

%To better understand what kind of annotations we need to analyze propaganda content, we propose to divide multimodal propaganda decryption into different sub-tasks to tackle different aspects of the problem and then design appropriate annotation scheme for each sub-task:% based on our investigation into the \textit{En Guardia} magazine:
% We model the propaganda messaging on four different levels:

\vspace{-2mm}
\paragraph{Task 1: Propaganda device and tactic detection and rationale generation.} 
The first task concerns identifying the propaganda device(s) and tactic(s) used in propaganda content. %following a well-known propaganda theory known as 
The "Seven Propaganda Devices" \cite{childs_1936, authorship}, a well-known propaganda theory, defines seven propaganda devices that represent the seven persuasion strategies commonly used in propaganda, including:
%that have become a gold standard in social science studies of propaganda. The seven devices are: 
\textit{Band Wagon}, \textit{Card Stacking}, \textit{Glittering Generalities}, \textit{Name-Calling}, \textit{Plain Folks}, \textit{Testimonial}, and \textit{Transfer}.%
\footnote{Other propaganda theories can also be used.}
We identify the device(s) used in the text portion and the image portion of the multimodal input separately. Note that more than one device can be used for a given piece of text or image.
%These devices are used to categorize individual messages in the text and images. While we anticipate that multiple devices may apply to a propaganda message, we rank the devices according to their prevalence in each message. Besides propaganda devices, 

In addition, we extend our annotation scheme by including a set of propaganda {\em tactics}, which serve to underscore the methods of employing the devices. The set of 
%propaganda 
tactics we have identified include: \textit{Extremism}, \textit{All-encompassing}, \textit{Repetition}, \textit{Visual scaling}, \textit{Binary Reduction}, \textit{Cultural Signaling}, \textit{Prestige Signaling}, \textit{Pandering}, and \textit{Innuendo}.

Finally, we provide the rationale behind each device label and each tactic label we assign to the given propaganda content. A rationale is a natural language description of why the corresponding device/tactic label is assigned based on the information extracted from the input. As we will see, rationale generation could improve model interpretability.

\vspace{-2mm}
\paragraph{Task 2: Domain-independent message detection with rationales.} Inspired by existing propaganda theory regarding the content of an article \cite{Ellul1973-na,9780205067169}, we detect the types of the (possibly hidden) messages conveyed by the author in our second task. The messaging in this task is domain-independent and attempts to communicate a broad idea to provoke an emotional reaction. An initial set of message types we identified includes: \textit{Might}, \textit{Fear}, \textit{Inspiration}, \textit{Urgency}, \textit{Unity}, \textit{Teamwork}, \textit{Patriotism}, \textit{Superiority}, \textit{Abundance}, \textit{Reciprocity}, \textit{Sacrifice}, \textit{Masculinity}, \textit{Ingenuity}, \textit{Virtue}, \textit{Progress}, \textit{Security}, \textit{Reassurance}, \textit{Fun}, and \textit{Sameness}.  We expect this list to grow as we identify additional types.
%Note that 
Multiple message types may be applicable to a given propaganda content. As in Task~1, in this task the rationale behind each annotated message will be annotated.

\vspace{-2mm}
\paragraph{Task 3: Domain-specific message detection with rationales.} Like Task 2, Task 3 also concerns identifying the types of messages conveyed by the author, but the message types in this task are domain-specific and therefore would need to be redesigned for each new domain. For wartime propaganda such as those that appear in \textit{En Guardia}, the message types could include: \textit{Military Strength}, \textit{Industrial Production}, \textit{US-Latin American Cooperation}, \textit{US culture}, \textit{US Leadership}, \textit{WW2}, \textit{Pan-Americanism}, \textit{War preparation}, \textit{Economic Interests}, \textit{Gendered messaging}, \textit{Civilian contributions}, and \textit{Common Culture}.
Again, multiple message types may be applicable to a given propaganda content.
% Even though the level III types are domain-specific, the idea of modeling can be brought to any domains.
As in Tasks~1 and~2, in this task the rationale behind each annotated message will be annotated.

\vspace{-2mm}
\paragraph{Task 4: Main message generation with rationales.} This task concerns generating the main message conveyed by the author in natural language. As in the first three tasks, here the rationale behind the main message will be annotated.
%This can be regarded as the "main points" of the author in natural language sentences.

\vspace{-2mm}
\paragraph{Task 5: Background knowledge.} As noted before, background 
%historical or cultural 
knowledge may be needed to properly interpret propaganda content. The background knowledge needed will be annotated in the form of natural language sentences.

\vspace{-2mm}
\paragraph{Task 6: Image captioning.} 
Existing image encoders may fail to encode all the details of an image, particularly when the image contains abstract concepts. 
To mitigate the difficulty of accurately extracting information from images, we propose an auxiliary task, image captioning, where we annotate the information present in an image in natural language so that the resulting caption is an equivalent textual representation of the image. 
With these annotations, we can train a model to first caption an image and use the resulting caption in lieu of the original image for further processing.
%\footnote{The reason why we do not simply use an off-the-shelf image captioning tool to perform Task~6 is that there is no guarantee that it will produce captions that contain the same level of details as the captions required for Task~6.}
%where a model needs to use natural language sentences to encode the information presented in images. The ultimate goal of image transcription is that using only the image transcription and the caption as input, the model achieves around the same performance as using the image and the caption as input.

%\paragraph{Appendant Sub-Task: Rationale} In sub-tasks 1, 2, and 3, a computer system needs to detect the boundaries and labels of some propaganda messaging. However, why a computer system assigns a label is completely unclear. Hence, we propose the sub-task rationale as an appendant sub-task to sub-tasks 1, 2, and 3, which asks a computer system to output its rationale in natural language sentences regarding why it assigns a specific label.

\subsection{Sample Annotation}

We propose to annotate each propagandistic article in the form of an {\em argument tree}, which is the representation used by argument mining researchers to represent the argumentative structure of a persuasive document \cite{stab-gurevych-2014-annotating}. In an argument tree, the root node corresponds to the main claim of the document, and each child of a node corresponds to a piece of supporting evidence (which can be a claim or a premise) for the parent. In other words, each edge denotes a support relation. A leaf node always corresponds to a premise, which by definition does not need any support.

To enable the reader to understand how to annotate a propagandistic article as an argument tree, we show in Figure~\ref{fig:exp} the argument tree that should be produced for input taken from Figure~2(b).
As we can see, the root node contains the main message (see Task~4). It has five children, which implies that it is supported by five pieces of evidence, including the domain-independent and domain-dependent messages derived from the image, the domain-independent and domain-dependent messages derived from the text and the rationale associated with the main message, which is also derived from the text. For each of the first four children, there are two child nodes, one corresponding to its rationale and the other corresponding to the device(s) and tactic(s) used. The fifth child, which is a rationale, is a leaf node. Note that a rationale always appears in a leaf node, the reason being that rationales are derived directly from either the image or the text (or both) and therefore do not need any support. The remaining nodes in the tree can be interpreted in a similar fashion. Note that the annotations for Tasks~1--5 will always appear as nodes in the tree.

\section{Models}

Given a dataset annotated using our annotation scheme, we can train a model to perform the six annotation tasks. Given the recent successes of neural models in NLP, we propose to employ neural models for our task. As a first step, we can employ existing models and design new models for this task if needed. There are several considerations.

\vspace{-2mm}
\paragraph{Multimodal vs. unimodal models.}
Since our input is multimodal and composed of text and image(s), it would be natural to train a multimodal model assuming three inputs: two of them correspond to the two modalities and the remaining one encodes the background knowledge base 
%\footnote{The background knowledge base could be 
(assembled using the background knowledge annotated for each training instance, for example). The images can be encoded using a visual encoder such as ResNet \cite{He2016DeepRL} and ViLBERT \cite{Lu2019ViLBERTPT}, whereas the text inputs (including the background knowledge base) could be encoded using a neural encoder such as SpanBERT \cite{joshi-etal-2020-spanbert}. The outputs from the encoders can then be concatenated together for further processing.

Alternatively, one can employ a unimodal model where we caption the image first (Task~6) with the help of an object detection system (e.g., YOLO \cite{yolo}) and possibly an off-the-shelf image captioning system (e.g., X-Transformer \cite{Pan_2020_CVPR}, VinVL \cite{Zhang_2021_CVPR}). As noted before, the caption is supposed to be an equivalent textual representation of the corresponding image. The caption can then be encoded by a text encoder, and the resulting representation can be concatenated with the encoded outputs from the text side for further processing.

\vspace{-2mm}
\paragraph{Joint vs. pipeline models.}
Should we adopt a pipeline architecture where we first train a model for each task independently of the others and then apply the resulting models in a pipeline fashion? For instance, given multimodal propagandistic articles, we can first train a model to caption the image (Task~6), as described above. After that, we can train a model to identify the device(s) and another model to identify the tactic(s) (Task 1). To improve model interpretability, the rationales can be predicted jointly with the corresponding device(s)/tactic(s).
%behind these device(s) and tactic(s) also need to be identified, we can train separate models to identify them, and the use the output of these models to help us predict the device(s) and tactic(s). 
We can similarly train models to predict the domain-independent and domain-specific labels (Tasks 2 and 3) jointly with their rationales by using
%by first training models to predict their rationales and then predict the labels by using the resulting rationales together with 
all of the available information predicted so far (e.g., the tactics and devices, the caption). Finally, we can train a model to predict the main message (Task 4).
%based on the caption and the information predicted so far.

Recall that pipeline models are prone to error propagation, where errors made by an upstream model will propagate to a downstream model. To mitigate error propagation, we can consider joint models. Specifically, we can train {\em one} model to perform all of the six tasks jointly. Joint models allow the different tasks involved to benefit each other via a shared input representation layer. The major downside of a joint model is that the resulting network (and hence the corresponding learning task) is typically very complex.

\vspace{-2mm}
\paragraph{Pre-trained models.} A key challenge in the automatic processing of propaganda is the need for background knowledge. While we have access to background knowledge through the manual annotations obtained as part of Task~5, it is conceivable that the amount of background knowledge we need will far exceed what these annotations can provide. A potential solution to this background knowledge acquisition bottleneck is {\em pre-training}.
More specifically, researchers in NLP have shown  that a vast amount of general knowledge about language, including both linguistic and commonsense knowledge, can be acquired by (pre-)training a language model in a {\em task-agnostic} manner using {\em self-supervised} learning tasks. Self-supervised learning tasks are NLP tasks for which the label associated with a training instance can be derived automatically from the text itself.%
\footnote{A well-known self-supervised learning task is Masked Language Modeling (MLM)~\cite{devlin2019bert}. Given a sequence of word tokens in which a certain percentage of tokens is {\em masked} randomly, the goal of MLM is to predict the masked tokens.}
%As can be easily imagined, a model for MLM can therefore be trained on instances where each one is composed of a partially masked sequence of word tokens and the associated ``class'' value is the masked tokens themselves.}
Because no human annotation is needed, a language model can be pre-trained on a large amount of labeled data that can be automatically generated, thereby acquiring a potentially vast amount of knowledge about language. Many pre-trained language models have been developed and widely used in NLP, such as BERT \cite{devlin2019bert}, XLNet \cite{yang2019xlnet}, RoBERTa \cite{liu2019roberta}, ELECTRA \cite{clark2019electra}, GPT-2 \cite{gpt2}, T5 \cite{raffel2020t5}, and BART \cite{lewis2020bart}. These models have been shown to offer considerable improvements on a variety of NLP tasks.

To acquire the background knowledge needed for processing the articles in {\em En Guardia}, for instance, we can pre-train a language model on as many {\em unannotated} articles in {\em En Guardia} as we can collect. The resulting model can then be optimized for a specific task by {\em fine-tuning} its parameters using the task-specific labeled data we obtained via our annotation process in the standard supervised fashion. While there has been a lot of work on developing pre-trained language models, the development of {\em multimodal} pre-trained models that can understand both text and images, which is what we need for multimodal propaganda processing, is a relatively unexplored area of research. %still in its infancy.

\comment{
\vspace{-2mm}
\paragraph{Deep vs. shallow understanding.} Since our ultimate goal is to train a model to predict the hidden message conveyed in the propaganda content (Task~4), we can take a {\em shallow} approach, where we train a model that takes the multimodal input and the background knowledge base and directly predicts the hidden message without predicting the results of the intermediate tasks (Tasks~1--4). Unlike the deep understanding approach where all the intermediate tasks will be learned, the shallow approach will not be interpretable (since no rationales will be outputted) and hence it will not learn to reason with and exploit the information provided by the intermediate tasks when predicting the hidden message. Nevertheless, the shallow approach could be sufficient as far as predicting the hidden message is concerned.
}

\section{Related Work}

\paragraph{Memes.} Memes are user-created pictures, often accompanied by text, that are used to express opinions on all kinds of topics. 
%Memes have been popular on every social media platform, thanks to their easy-to-make, easy-to-understand, and easy-to-spread natures. 
Similar to propaganda messaging, memes typically require 
%a human reader to have certain 
background knowledge for proper interpretation. %At the same time, 
Memes are widely used in political manipulation campaigns as a tool for conveying propaganda messaging \cite{farwell2014media,fhms15,bradshaw2018challenging,newKnowledge}. Hence, unveiling hidden information from memes is highly related to processing propaganda messaging from images and text. There has been recent work that aims to build a model to detect a rich set of propaganda techniques in memes \cite{dimitrov-etal-2021-detecting}.

\vspace{-2mm}
\paragraph{Document-level unimodal misinformation analysis.} Several publicly-available datasets % have been released where 
are composed of news articles labeled with whether they contain misinformation. 
%Most research on propaganda identification in that past considers only textual information on a quite limited label space \cite{rashkin-etal-2017-truth, 10.1016/j.ipm.2019.03.005}. 
For example, in the \texttt{TSHP-17} dataset \cite{rashkin-etal-2017-truth}, each article is labeled with one of four classes: \textit{trusted}, \textit{satire}, \textit{hoax}, and \textit{propaganda}, whereas in the \texttt{QProp} dataset \cite{10.1016/j.ipm.2019.03.005}, only two labels are used: \textit{propaganda} and \textit{non-propaganda}.
\citet{da-san-martino-etal-2019-fine}, on the other hand, develop a corpus of news articles %that contains a rich set of 18
%each of which is 
labeled with the propaganda techniques used. Their corpus enables the study of multi-label multi-class classification task in propaganda identification. 
%The dataset is also used in the shared task on Fine-grained Propaganda Detection @NLP4IF 2019 and SemEval-2020 Task 11 on Detection of Propaganda Techniques in News Articles \cite{da-san-martino-etal-2019-findings,da-san-martino-etal-2020-semeval}. 

\vspace{-2mm}
\paragraph{Multimodal misinformation classification.} 
%Thanks to the recent advances of visual encoders and pretrained language models, 
%Recently, 
Some researchers have examined the task of multimodal propaganda identification.
%where the goal is to classify a multmodal social media post consisting of misleading information into a predefined set of categories. 
For instance, \citet{Volkova_Ayton_Arendt_Huang_Hutchinson_2019} construct a dataset consisting of 500,000 Twitter posts classified into six categories (\textit{disinformation}, \textit{propaganda}, \textit{hoaxes}, \textit{conspiracies}, \textit{clickbait}, and \textit{satire}) and build models to detect misleading information in images and text. %\citet{10.5555/3495724.3495944} proposed a dataset for detecting hateful speech in multimodal memes. \citet{1909.05838}. 
While this attempt seeks to perform a shallow analysis of tweets, we propose to perform a deep analysis of propaganda content, which would lead to the generation of the hidden messages conveyed.
%explored different multilingual models for detecting deception in the multimodal setting.

%Our work differs from the above mentioned works in three aspects. First, we develop an annotation scheme that considers multiple layers of propaganda messaging including propaganda techniques, domain-specific and domain-independent propaganda messaging. Second, differ from past research which usually focuses on a given piece of text/image, we consider propaganda identification on a brand new level --- the sequence level, where a sequence of text/images need to be considered. Third, when textual information is considered, oftentimes the rich formatting is excluded from consideration. This causes a great loss of propaganda information, as human readers would naturally be attracted to the rich formatting of text. In our annotation scheme, we propose that this rich formatting information should be kept in annotations.

\section{Concluding Remarks}

We presented the task of multimodal propaganda processing and discussed the key challenges.
%and outline the first steps involved in addressing this task. 
Below we conclude with several other
issues that %we have eluded so far but which 
are also relevant to the task.

\vspace{-2mm}
\paragraph{Propaganda identification.} While we have focused on analyzing propaganda content, 
%in this paper, 
it is equally important to %automatically 
identify such content. Although we did not explicitly discuss how such content can be identified, %we believe that 
a system that can analyze propaganda content could also be used for identifying such content. More specifically, if the system determines that no persuasion devices and tactics were being used in the given content, it could imply that the content is not propagandistic. Another possibility would be to train a model to distinguish propaganda content from non-propaganda content on our corpus of propaganda articles and other non-propaganda articles  collected from the Internet.

\vspace{-2mm}
\paragraph{Domain transferability.} 
Since the models described thus far are trained on domain-specific annotations (i.e., the background knowledge from Task~5 and the domain-specific labels from Task~4), they are necessarily domain-specific. To facilitate their application to a new domain, especially when labeled training data in the new domain is scarce, we can explore domain adaptation techniques.
%by, for instance, mixing the training data from the original domain and those from the target domain for model training.

\vspace{-2mm}
\paragraph{Ethical considerations.} 
%Although it will likely take years or even decades to fully address this task, 
%We as 
%AI researchers should be aware of the ethical considerations associated with the development of propaganda processing systems. For instance, 
Care should be taken to ensure that %this kind of 
propaganda processing technologies would not be misused by people to attack their political opponents by intentionally using a propaganda processing system to draw wrong conclusions or generate propaganda content aiming to achieve their personal agenda, for instance.

%where the goal is to analyze propaganda content by extracting relevant information from the different modalities present in the content, identifying the persuasion devices and tactics that were used in different portions of the content, and eventually generating the message(s) being conveyed. We discussed the challenges this task presents to AI researchers. Despite these challenges, we believe that time is ripe to start examining this task, particularly in light of the increasing prevalence of propaganda content on the Web. The subtasks that we identified and the corpus construction procedure that we outlined should provide the first steps towards addressing this task. 
%While it is important to analyze multimodal propaganda content, 

%\section*{Acknowledgments}
%We thank the three anonymous reviewers for their
%detailed and 
%insightful 
%comments on an earlier
%draft of the paper. 
%This work was supported in part by NSF Grant IIS-1528037.
%Any opinions, findings, conclusions or recommendations expressed in this paper are those of the authors and do not necessarily reflect the views or official policies, either expressed or implied, of the NSF. 

% Use \bibliography{yourbibfile} instead or the References section will not appear in your paper
\bibliography{aaai23,new}

\end{document}